\documentclass[letterpaper,10 pt,conference]{ieeeconf}
\IEEEoverridecommandlockouts
\usepackage{float}
\usepackage[dvipdfmx]{graphicx}
\usepackage{amsmath}
\usepackage{amssymb}
\usepackage[space]{cite}
\usepackage{bm}
\usepackage{booktabs}
\usepackage{tabularx}
\usepackage{url}
\usepackage{multirow}
\usepackage{gensymb}
\usepackage{dsfont}
\usepackage[x11names, rgb]{xcolor}
\usepackage{colortbl}
\usepackage{url}
\usepackage{siunitx}
\usepackage{threeparttable}
\usepackage{graphbox}
\usepackage{tikz}

\usepackage{adjustbox}
\usepackage{mwe} 

\usepackage{siunitx}
\usepackage{pifont}
\newcommand{\cmark}{\ding{51}}%
\newcommand{\xmark}{\ding{55}}%

\usepackage{tikz}
\usetikzlibrary{backgrounds}
\newcommand{\twocolorsquare}[2]{
    \begin{tikzpicture}[baseline=0.8pt,scale=0.9]
        \fill[#1] (0, 0) -- (7pt, 7pt) -- (0, 7pt) -- cycle;
        \fill[#2] (7pt, 0) -- (7pt, 7pt) -- (0, 0) -- cycle;
    \end{tikzpicture}%
}

\makeatletter
\let\NAT@parse\undefined
\makeatother
\usepackage{hyperref}
\hypersetup{
	bookmarks         = true,
	bookmarkstype     = toc,
	bookmarksnumbered = true,
	bookmarksopen     = true,
	colorlinks        = false,
	hidelinks,
	breaklinks        = false,
}

\usepackage[capitalize]{cleveref}
\crefname{section}{Sec.}{Secs.}
\Crefname{section}{Section}{Sections}
\Crefname{table}{Table}{Tables}
\crefname{table}{Tab.}{Tabs.}

\newcolumntype{C}{>{\centering\arraybackslash}X}
\newcolumntype{L}{>{\raggedright\arraybackslash}X}
\newcolumntype{R}{>{\raggedleft\arraybackslash}X}
\newcolumntype{P}[1]{>{\centering\arraybackslash}p{#1}}

\newcommand{\etal}{\textit{et al.}}

\newcommand{\s}{\hphantom{0}}

\newcommand{\mysquare}[1][black]{{\small\textcolor[rgb]{#1}{\ensuremath\blacksquare}}}
\newcommand{\myparagraph}[1]{\noindent\textbf{#1}}

\title{\LARGE \bf
DRUM: Diffusion-based Raydrop-aware Unpaired Mapping \\for Sim2Real LiDAR Segmentation
}

\author{Tomoya Miyawaki$^1$ \quad Kazuto Nakashima$^2$ \quad Yumi Iwashita$^3$ \quad Ryo Kurazume$^2$%
\thanks{*This work was supported by JSPS KAKENHI Grant Number JP23K16974 and JP20H00230.}%
\thanks{$^{1}$Tomoya Miyawaki is with the Graduate School of Information Science and Electrical Engineering, Kyushu University, Japan. {\tt\small miyawaki@irvs.ait.kyushu-u.ac.jp}}%
\thanks{$^{2}$Kazuto Nakashima and Ryo Kurazume are with the Faculty of Information Science and Electrical Engineering, Kyushu University, Japan. {\tt\small k\_nakashima@mech.kyushu-u.ac.jp, kurazume@ait.kyushu-u.ac.jp}}%
\thanks{$^{3}$Yumi Iwashita is with the Jet Propulsion Laboratory, California Institute of Technology, USA. {\tt\small yumi.iwashita@jpl.nasa.gov}}}

\begin{document}

\maketitle
\thispagestyle{empty}
\pagestyle{empty}

\begin{abstract}
	LiDAR-based semantic segmentation is a key component for autonomous mobile robots, yet large-scale annotation of LiDAR point clouds is prohibitively expensive and time-consuming. Although simulators can provide labeled synthetic data, models trained on synthetic data often underperform on real-world data due to a data-level domain gap. To address this issue, we propose DRUM, a novel Sim2Real translation framework. We leverage a diffusion model pre-trained on unlabeled real-world data as a generative prior and translate synthetic data by reproducing two key measurement characteristics: reflectance intensity and raydrop noise. To improve sample fidelity, we introduce a raydrop-aware masked guidance mechanism that selectively enforces consistency with the input synthetic data while preserving realistic raydrop noise induced by the diffusion prior. Experimental results demonstrate that DRUM consistently improves Sim2Real performance across multiple representations of LiDAR data. The project page is available at \url{https://miya-tomoya.github.io/drum}.
\end{abstract}

\section{Introduction}
\label{sec:introduction}

3D LiDAR sensors capture high-fidelity point clouds of the surrounding environment using time-of-flight (ToF) ranging.
These LiDAR point clouds have become integral to scene perception for autonomous systems, such as mobile robots and self-driving cars.
In particular, semantic segmentation of LiDAR point clouds has been a central task in robotics and computer vision research~\cite{milioto2019rangenet++,hu2020randla-net,wu2019squeezesegv2,zhao2021epointda,choy20194d}, which involves assigning a semantic label to every point in a scene.
However, the dominant supervised learning paradigm for this task relies heavily on large quantities of \emph{labeled} point clouds, which are prohibitively expensive in terms of manual labor and time.
For instance, annotating the widely used SemanticKITTI benchmark~\cite{behley2019semantickitti} required over 1,700 person-hours for its 43,000 scans.
This annotation bottleneck motivates research into cost-effective alternatives.

\begin{figure}[t]
	\centering
	\scriptsize
	\includegraphics[width=1.0\hsize]{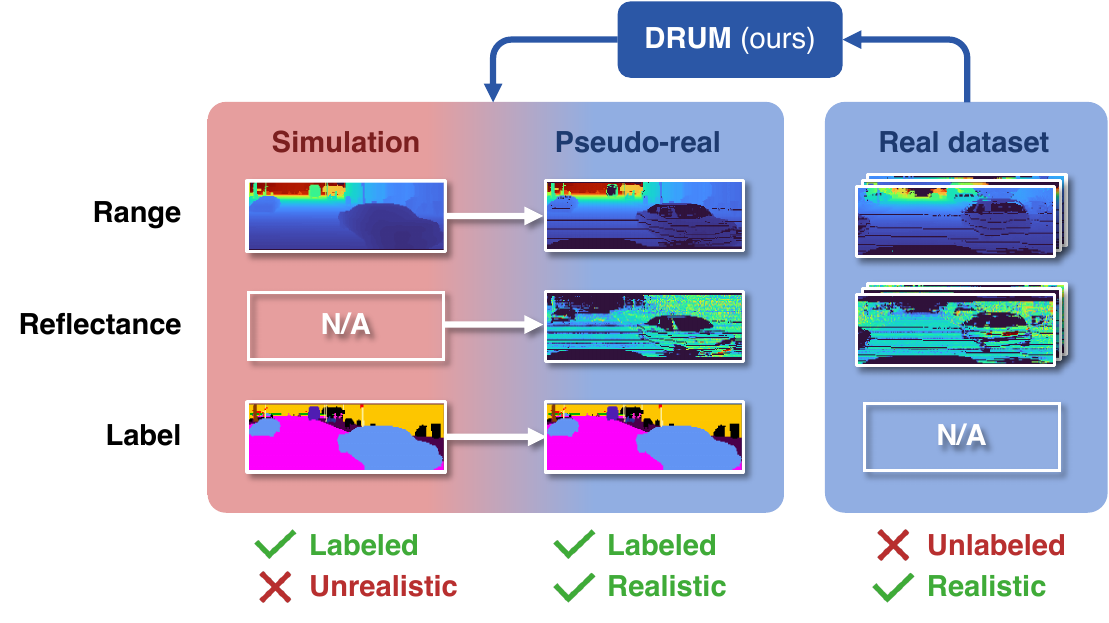}
	\vspace*{-5mm}
	\caption{\textbf{Problem formulation}. Given unpaired sets of labeled simulation and unlabeled real samples, our framework DRUM generates labeled pseudo-real samples for training LiDAR segmentation.}
	\label{fig:teaser}
\end{figure}

Sim2Real (simulation-to-real) transfer~\cite{wu2018squeezeseg,wu2019squeezesegv2,zhao2021epointda,nakashima2023generative} offers a promising solution to the critical annotation bottleneck in LiDAR-based perception.
This paradigm leverages simulators equipped with ray-tracing and physics engines to generate vast quantities of automatically labeled synthetic data, circumventing the need for laborious manual annotation.
However, models trained exclusively on synthetic data often fail to generalize to real-world data due to the inherent domain gap between simulation and real domains.
A major source of this gap stems from two factors governed by both scene properties and sensor characteristics: \emph{reflectance intensity} and \emph{raydrop noise}.
Reflectance intensity is the measured strength of the backscattered laser signal, influenced by factors such as surface material, incidence angle, and sensor characteristics.
Raydrop noise occurs when the laser signal attenuates or scatters upon hitting a surface, resulting in an insufficient return intensity for detection.
It appears as missing pixels in range or reflectance images, as exemplified in~\cref{fig:teaser}.
Although existing LiDAR simulators~\cite{dosovitskiy2017carla,isaacsim,jansen2023cosys-airsim} implement simplified physics models to simulate these effects, they still fall short of capturing the complexity and diversity observed in real-world measurements.

To bridge the domain gap between simulation and the real world, there has been a growing interest in developing machine learning systems to simulate realistic LiDAR data.
One prominent line of work is \emph{re-simulation}~\cite{manivasagam2020lidarsim,zhang2024nerf-lidar}, which synthesizes novel sensor views from an existing set of real-world scans.
While this method allows for the reuse of initial annotations, it lacks scalability as it still relies on a manually labeled real-world dataset.
Another approach focuses on \emph{Sim2Real translation}, converting labeled synthetic data to the real-world domain.
This is often accomplished by building conversion models through self-supervised learning that leverages superimposed multi-scan observations~\cite{guillard2022learning,zyrianov2025lidardm} or unsupervised learning with deep generative models~\cite{zhao2021epointda,nakashima2023generative,xiao2022transfer}, such as generative adversarial networks (GANs)~\cite{goodfellow2014generative}.
Nevertheless, these methods still struggle to jointly and realistically reproduce the complex patterns of raydrop and reflectance intensity.

To this end, we propose \textbf{DRUM} (diffusion-based raydrop-aware unpaired mapping), a novel Sim2Real translation method for semantic segmentation of LiDAR point clouds.
The problem formulation is illustrated in~\cref{fig:teaser}.
Our method formulates the translation task as posterior sampling of real-world range and reflectance data, conditioned only on the range data from the simulation domain.
To represent the prior distribution of the real-world domain, we use diffusion models~\cite{ho2020denoising}, a family of deep generative models that have garnered significant attention in recent years for their high-fidelity generation capabilities.
Although posterior sampling using pre-trained diffusion models has been demonstrated in general restoration tasks~\cite{song2023pseudoinverse-guided} including a LiDAR upsampling task~\cite{nakashima2024lidar,zyrianov2022learning}, we found that inconsistency in raydrop noise between the simulation and real domains causes unstable and unrealistic generation.
To address this issue, we introduce two key techniques: a raydrop-aware masked sampling strategy based on a provisional estimate during sampling, and an initialization technique to ensure geometric consistency.

Our contributions can be summarized as follows:
\begin{itemize}
	\item We propose a novel Sim2Real translation method for LiDAR point clouds. We formulate the task as posterior sampling using unconditional diffusion models pre-trained on the real-world dataset, enabling a unified generation of realistic range, reflectance, and raydrop.
	\item We propose a raydrop-aware masked guidance mechanism that conditions the generation process on simulation data, thereby stabilizing the diffusion-based posterior sampling.
	\item We demonstrate that our approach outperforms existing methods in terms of sample fidelity and Sim2Real performance in semantic segmentation. 
\end{itemize}

\section{Related Work}
\label{sec:related_work}

\textbf{Semantic segmentation of LiDAR data} is a task of predicting point-level semantic classes, such as cars and roads in driving scenes~\cite{behley2019semantickitti}.
Various segmentation models have been proposed primarily using neural networks, depending on the representation of LiDAR data, such as point clouds~\cite{hu2020randla-net}, images~\cite{milioto2019rangenet++,wu2019squeezesegv2}, and voxels~\cite{choy20194d}.
In many cases, multimodal information, \textit{i.e.}, the measured range and reflectance intensity, is used as input cues for semantic segmentation.
Manual annotation is required to construct training data for neural networks, which entails substantial cost and time, as it requires consistent quality for the massive number of 3D points.

\textbf{Sim2Real transfer} is one approach to solving the annotation problem.
A variety of simulators leveraging graphics and physics engines, such as CARLA~\cite{dosovitskiy2017carla}, are available, which enable us to automatically collect training data with high-quality annotations.
Leveraging these simulators, a number of datasets have been introduced~\cite{wu2019squeezesegv2,xiao2022transfer}, featuring diverse sensor setups and object categories.
However, their generalization performance inevitably suffers from a domain gap between simulation and real-world samples.
To address these issues, Sim2Real transfer is commonly treated as a domain adaptation problem.
Existing approaches are twofold: feature-level mapping and data-level mapping.
The feature-level mapping aims to align the distributions between simulation and real domains in a shared feature space, such as through domain-invariant feature learning~\cite{ganin2016domain-adversarial}.
The data-level mapping approaches attempt to transform synthetic data to appear more realistic, such as by reproducing raydrop noise~\cite{wu2019squeezesegv2,zhao2021epointda} and reflectance intensity~\cite{xiao2022transfer}.
Nevertheless, challenges persist with respect to the fidelity of the reproduced modalities and the geometric consistency of the scenes.
Furthermore, most prior work addresses these problems in isolation, despite their multimodal nature.
We propose a unified framework to estimate the raydrop and reflectance intensity, leveraging the following deep generative models.

\textbf{Deep generative models} have recently emerged as a powerful framework for capturing the distributions of a wide variety of data, including images and videos.
These models can synthesize novel data by sampling from a learned distribution.
Furthermore, the trained model can be used as a data prior for downstream tasks such as restoration and editing.
Numerous generative methods have also been developed for the LiDAR domain~\cite{caccia2019deep,xiong2023learning,nakashima2021learning,nakashima2023generative,zyrianov2022learning,nakashima2024lidar,nakashima2025fast,zyrianov2025lidardm,ran2024towards} based on variational autoencoders (VAEs)~\cite{kingma2021variational}, generative adversarial networks (GANs)~\cite{goodfellow2014generative}, and diffusion models~\cite{ho2020denoising,song2021score-based}.
In particular, approaches using diffusion models~\cite{zyrianov2022learning,nakashima2024lidar,nakashima2025fast,zyrianov2025lidardm,ran2024towards} have significantly improved the quality of generated data and the stability of model training against other approaches.
Diffusion models describe the data generation process using stochastic or ordinary differential equations (SDE/ODE), which are guided by the gradient of the time-evolving data distribution, known as the \emph{score}.
This formulation allows the models to be adapted for conditional generation by modifying the score via Bayesian inference, which has led to their application in downstream tasks such as upsampling and completion.
In this paper, we repurpose the power of diffusion models for the task of Sim2Real domain adaptation of LiDAR data.

\section{Method}
\label{sec:method}

\subsection{Problem Formulation}
\label{sec:problem_formulation}

Given an unpaired set consisting of a \emph{labeled} simulation dataset $\mathcal{D}_{\mathrm{sim}}$ and an \emph{unlabeled} real dataset $\mathcal{D}_{\mathrm{real}}$, we aim to address the problem of unpaired Sim2Real translation, as illustrated in~\cref{fig:teaser}.
More specifically, $\mathcal{D}_{\mathrm{sim}}$ contains only the \emph{range} modality, whereas $\mathcal{D}_{\mathrm{real}}$ contains both \emph{range} and \emph{reflectance} modalities.
Furthermore, the real samples exhibit raydrop noise, which manifests as missing pixels in the images.

\myparagraph{Data representation.}
Since the majority of LiDAR generative models~\cite{caccia2019deep,nakashima2021learning,zyrianov2022learning,nakashima2023generative,nakashima2024lidar,ran2024towards,nakashima2025fast} have been developed based on the image representation, we also perform the data translation on the same image space.
We assume a spinning multi-beam LiDAR with angular resolution $W$ in azimuth and $H$ in elevation that measures range and reflectance at each angle $(\theta,\phi)$.
These $H\!\times\!W$ measurements are projected onto a two-channel equirectangular image space $\mathbb{R}^{2\times H \times W}$.
The simulation data contain only range modality, and the reflectance channel is zero-padded.
Here, we first formulate the absence of the reflectance channel as a linear degradation of the sample from the real domain to the simulation domain, and then consider the following linear inverse problem to recover a real-domain sample.

\myparagraph{Linear inverse problem.}
A general form of a linear degradation process can be written as the following observation equation:
\begin{equation}
	\label{eq:observation_equation}
	\bm{y} = H \bm{x} + \bm{n},
\end{equation}
where $\bm{y}\in\mathbb{R}^{m}$ is an observation, $H \in \mathbb{R}^{m \times n}$ is a known measurement matrix, $\bm{x} \in \mathbb{R}^n$ is an original signal, and $\bm{n} \in \mathbb{R}^m$ is the observation noise drawn from $\mathcal{N}(0, \sigma^2\bm{I})$.
The goal of the linear inverse problem is to estimate $\bm{x}$ from the observed data $\bm{y}$.

\myparagraph{Our approach in LiDAR Sim2Real.}
Instantiating $H$ as a corruption matrix that zeroes the reflectance channel, our task can be defined as estimation of $\bm{x}$ in real domain that contains both range and reflectance from $\bm{y}$ in simulation domain that contains only range information.
We note that, while this formulation defines the \emph{forward} corruption of reflectance on $\bm{y}\leftarrow\bm{x}$, it cannot account for the non-linear effects of raydrop noise, \textit{i.e.}, \emph{backward} corruption on $\bm{y}\rightarrow\bm{x}$.
Therefore, we formulate Sim2Real translation as a hybrid inference problem: explicitly solving a linear inverse problem for reflectance restoration with the known $H$, while implicitly modeling the non-linear raydrop degradation through our proposed mechanism, as detailed in~\cref{sec:progressive_masking}.

\subsection{Diffusion Posterior Sampling}
In this section, we describe a method for solving the aforementioned linear inverse problem through posterior sampling $\bm{x} \sim p(\bm{x}\mid\bm{y})$ with diffusion models.
We first review the unconditional sampling using a diffusion model~\cite{song2021score-based}, describe how we can solve the general linear inverse problems~\cite{song2023pseudoinverse-guided} by leveraging the same pre-trained diffusion model, and then identify the challenges in our case.

\myparagraph{Unconditional sampling.}
Diffusion models formulate an unconditional sampling of the data distribution $\bm{x} \sim p(\bm{x})$ based on the following reverse-time SDE~\cite{song2021score-based}:
\begin{equation}
	\label{eq:reverse_sde}
	d\bm{x}_t = \left[ f(\bm{x}_t,t) - g^2(t)\nabla_{\bm{x}_t}\log{p(\bm{x}_t)} \right] dt + g(t)\,d\bar{\bm{w}},
\end{equation}
where the state variable $\bm{x}_t$ evolves over the continuous time $t\in[1,0]$, $f(\bm{x}_t,t)$ and $g(t)$ are coefficient functions designed according to the types of diffusion models, $\nabla_{\bm{x}_t} \log{p(\bm{x}_t)}$ is a learnable score function modeled by a neural network, and $\bar{\bm{w}}$ is the standard Wiener process.
We can sample data by integrating~\cref{eq:reverse_sde}, initializing with a Gaussian sample at $t=1$.
One can approximate~\cref{eq:reverse_sde} by deterministic samplers such as probability flow ODE~\cite{song2021score-based} and DDIM~\cite{song2021denoising}.

\myparagraph{Posterior sampling.}
Linear inverse problems can also be addressed within a similar framework.
To sample $\bm{x}$ conditioned on $\bm{y}$, we aim to draw samples from the posterior distribution $p(\bm{x}\mid\bm{y})$.
To this end, we can replace the score in~\cref{eq:reverse_sde} with the following conditional score:
\begin{equation}
	\label{eq:conditional Score}
	\nabla_{\bm{x}_t} \log{p(\bm{x}_t\mid\bm{y})} = \underbrace{\nabla_{\bm{x}_t} \log{p(\bm{x}_t)}}_{\text{prior}} + \underbrace{\nabla_{\bm{x}_t} \log{p(\bm{y}\mid\bm{x}_t)}}_{\text{guidance}},
\end{equation}
where we use the Bayes' rule: $p(\bm{x}\mid\bm{y}) \propto p(\bm{x}) p(\bm{y}\mid\bm{x})$.
It is worth noting that the first \emph{prior} term is the unconditional score that can be trained solely $\bm{x}$ based on~\cref{eq:reverse_sde}.
Moreover, the second \emph{likelihood} term is referred to as \emph{guidance}, which can be approximated from~\cref{eq:observation_equation}.
We employ the definition of $\Pi$GDM~\cite{song2023pseudoinverse-guided} as follows:
\begin{equation}
	\label{eq:pigdm}
	\nabla_{\bm{x}_t} \log{p(\bm{y}\mid\bm{x}_t)} \approx r_t^{-2} \left[ \left( H^{\dagger} \bm{y} - H^{\dagger} H \hat{\bm{x}}_t \right)^\top \frac{\partial \hat{\bm{x}}_t}{\partial \bm{x}_t} \right]^\top,
\end{equation}
where $\hat{\bm{x}}_t$ is a tentative estimate of $\bm{x}_0$ given $\bm{x}_t$ using Tweedie's formula~\cite{stein1981estimation}, $H^{\dagger}$ is a Moore-Penrose pseudoinverse matrix of $H$, and $r_t$ is a time-dependent standard deviation that scales the gradient according to the noise level at timestep $t$.

\myparagraph{Challenges.}
Based on the above formulation, the pseudoinverse guidance in~\cref{eq:pigdm} computes a likelihood gradient at each time step, driven by the sim–real discrepancy $H^{\dagger}\bm{y}-H^{\dagger}H\hat{\bm{x}}_t$.
This guidance enforces strong consistency with the observation $\bm{y}$ and promotes the inpainting of missing content, ensuring that the samples conform to the observation.
However, applying it directly in our task is problematic.
Uniformly evaluating the discrepancy across all pixels drives the reconstruction $\hat{\bm{x}}_t$ to replicate the clean simulation $\bm{y}$.
Since $\bm{y}$ contains no raydrops, the guidance suppresses the realistic drop patterns that the diffusion model begins to form, resulting in ``over-inpainted'' and degraded samples.

\subsection{Raydrop-aware Masked Guidance}
\begin{figure*}[t]
	\centering
	\footnotesize
	\includegraphics[width=\hsize]{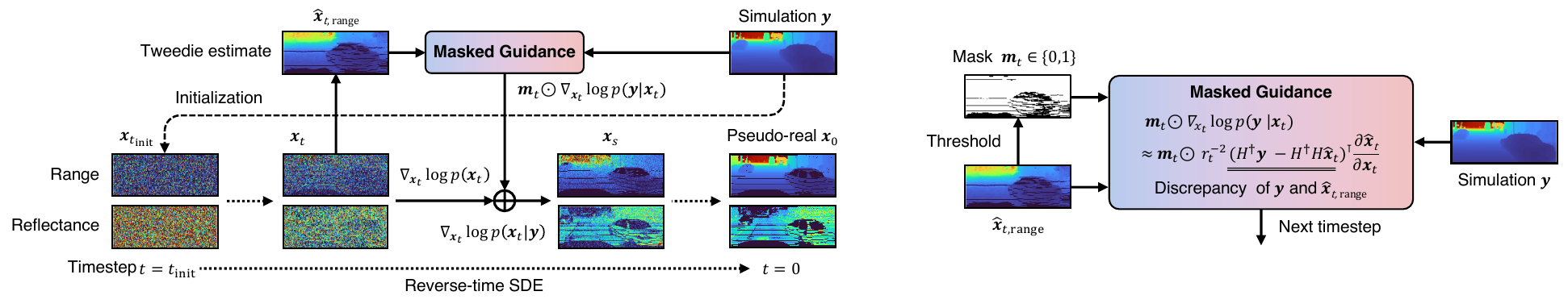}\\
	\begin{tabularx}{\hsize}{@{}m{0.5\hsize}@{\hspace{3mm}}m{0.5\hsize}@{}}
		\centering (a) Posterior sampling using a pre-trained diffusion model &   
		\centering (b) Masked guidance to condition with the simulation range $\bm{y}$
	\end{tabularx}
	\caption{
		\textbf{Overview of Sim2Real translation by DRUM}. 
		(a) The unconditional generation process of diffusion models is conditioned by the masked guidance with the simulation sample $\bm{y}$. 
		(b) In masked guidance, we first generate the raydrop-aware $\bm{m}_t$ mask from the tentative Tweedie sample $\hat{\bm{x}}_t$ and then compute the sim–real discrepancy based on the pseudoinverse method~\cite{song2023pseudoinverse-guided}. 
		The operator $H$ corrupts the reflectance modality.
	}
	\label{fig:model}
\end{figure*}

\begin{figure}[t]
	\centering
	\footnotesize
	\setlength{\tabcolsep}{2pt}
	\begin{tabularx}{\hsize}{C|CC}
		\includegraphics[align=c,width=\hsize]{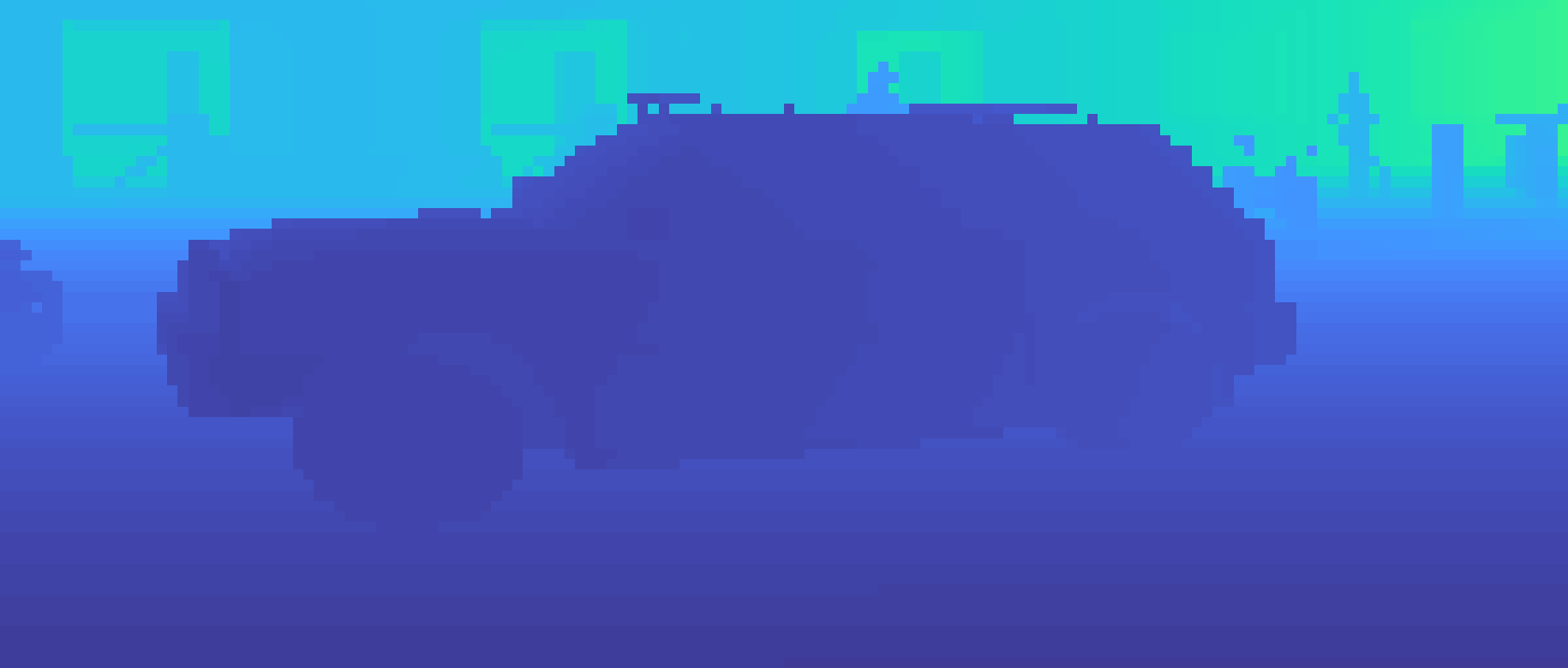} & 
		\includegraphics[align=c,width=\hsize]{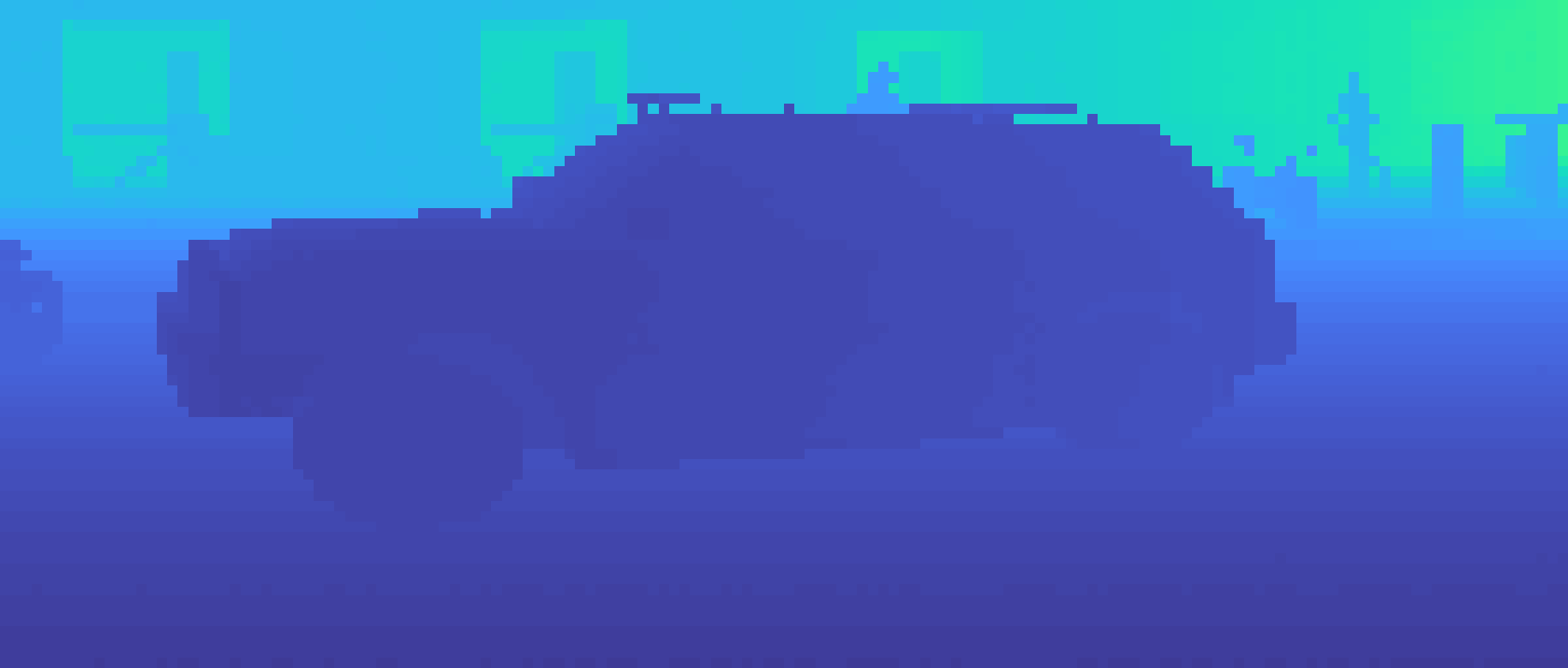} & 
		\includegraphics[align=c,width=\hsize]{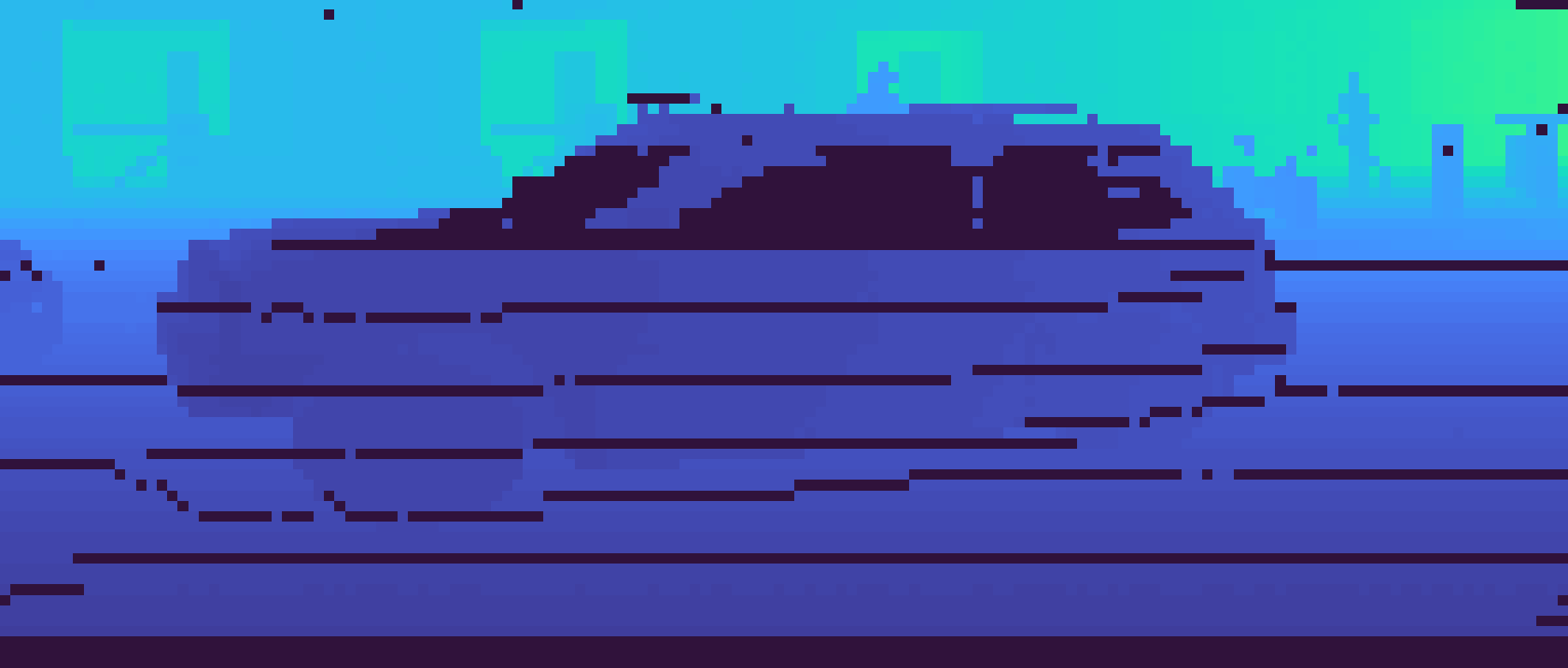} \\[15pt]
		\includegraphics[align=c,width=\hsize]{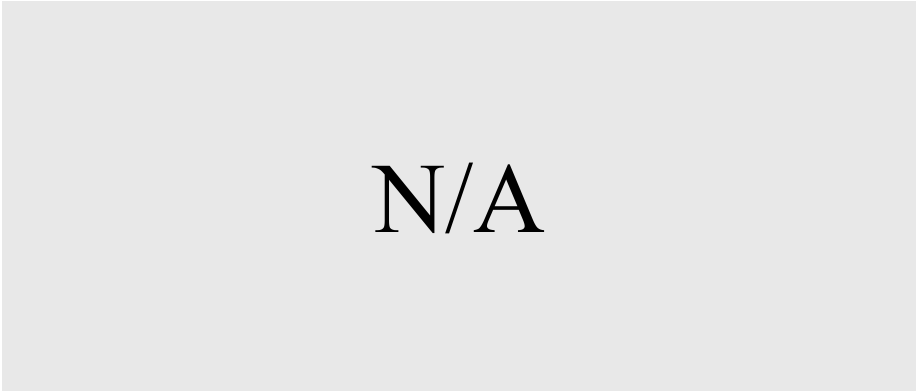} & 
		\includegraphics[align=c,width=\hsize]{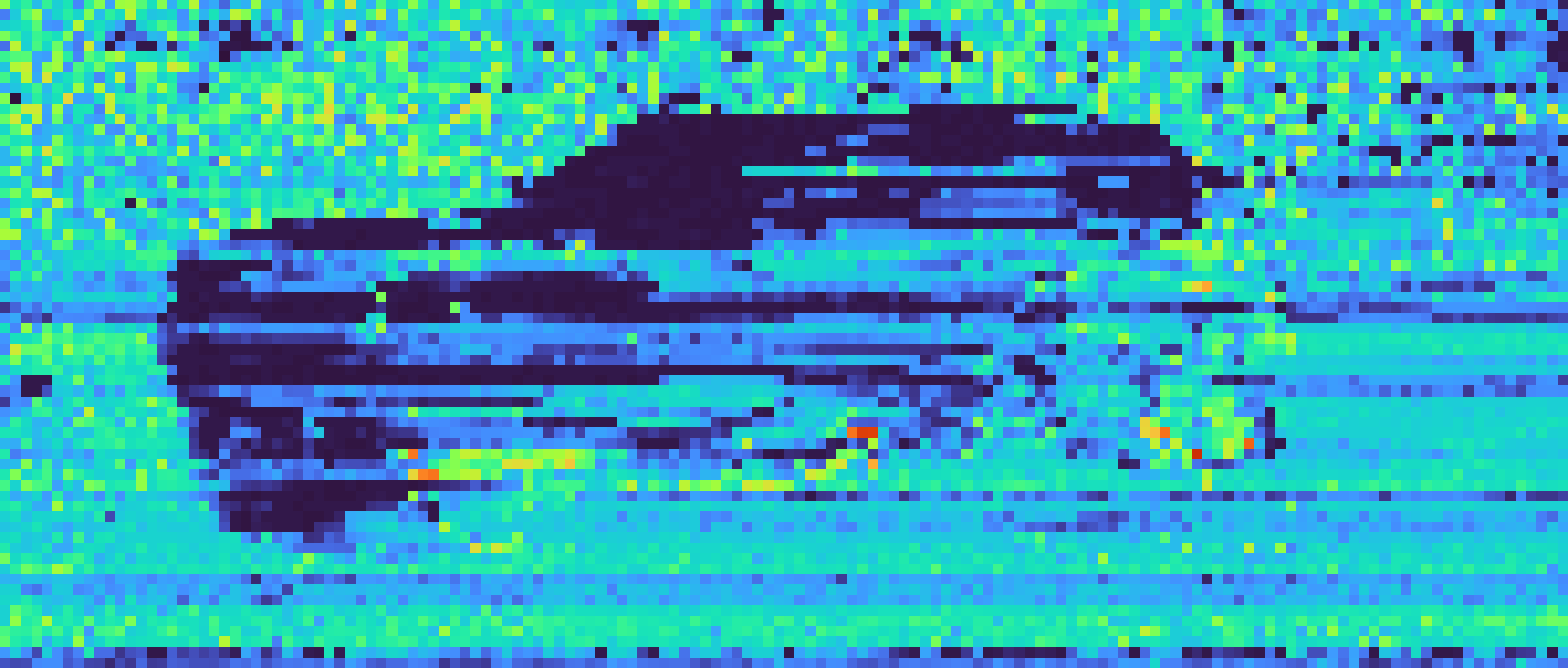} & 
		\includegraphics[align=c,width=\hsize]{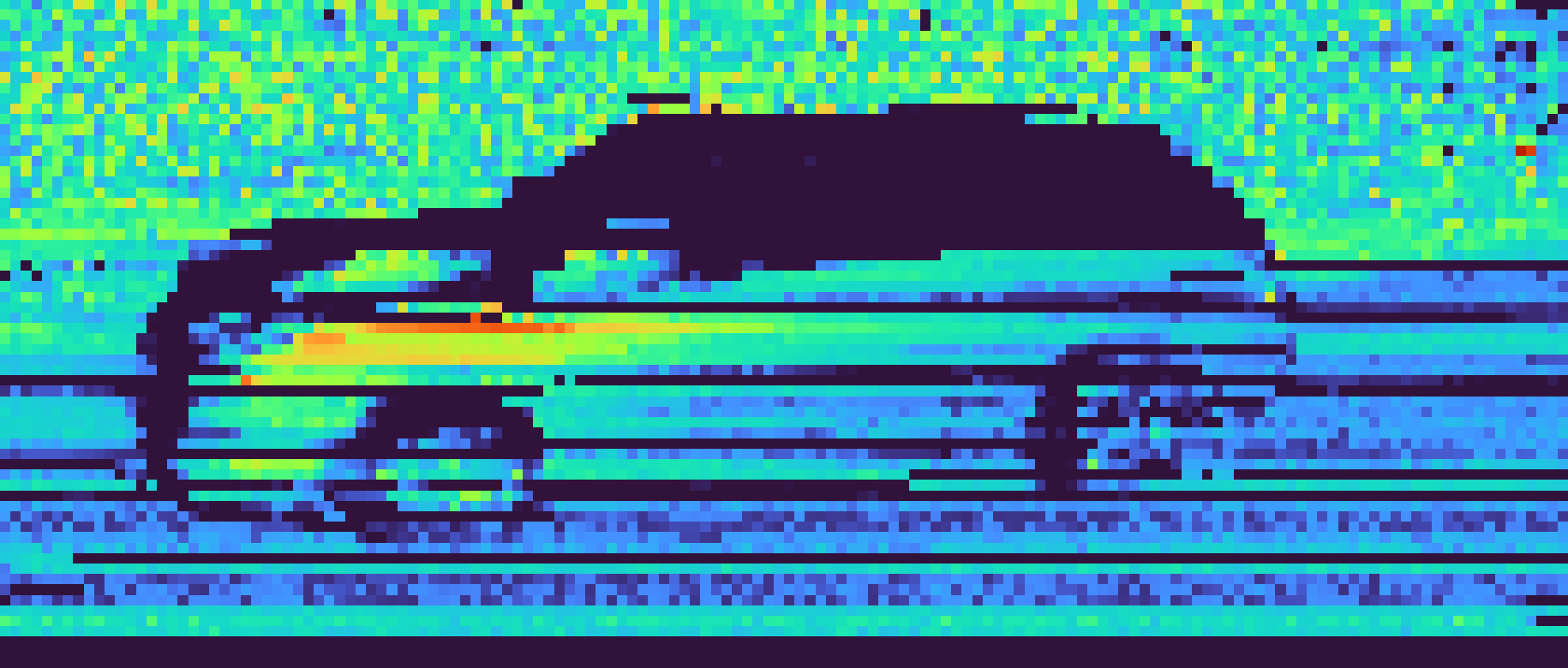} \\[15pt]
		Input        & Output \textbf{w/o ours} & Output \textbf{w/ ours} \\
		(simulation) & (pseudo-real)            & (pseudo-real)           
	\end{tabularx}
	\caption{\textbf{Ablation of our masked guidance}. We compare the range (top row) and reflectance (bottom row) pseudo-real samples produced with and without our raydrop-aware masked guidance. Our method successfully reproduces the raydrop noise on the car, as well as the reflectance modality.}
	\label{fig:ablation_masked_guidance}
\end{figure}

To avoid the above failure mode, we selectively gate the guidance in sampling, so that the raydrop pixels induced by the prior term are not overwritten by the simulation input $\bm{y}$.
\cref{fig:model} illustrates the overall framework of our posterior sampling and \cref{fig:ablation_masked_guidance} shows the generated samples.

\myparagraph{Progressive masking.}
\label{sec:progressive_masking}
The core idea is to modulate the guidance term of~\cref{eq:conditional Score} with a \emph{raydrop-aware mask} $\bm{m}_t$ estimated at each timestep $t$ as follows:
\begin{equation}
	\label{eq:masked_pigdm}
	\nabla_{\bm{x}_t} \log{p(\bm{x}_t\mid\bm{y})}=\nabla_{\bm{x}_t} \log{p(\bm{x}_t)} + \bm{m}_t \odot \nabla_{\bm{x}_t} \log{p(\bm{y}\mid\bm{x}_t)},
\end{equation}
where $\odot$ denotes the element-wise product.
Similar to the guidance computation, we leverage the Tweedie estimate $\hat{\bm{x}}_t$ to construct $\bm{m}_t$ at each timestep $t$.
Since the diffusion model is pre-trained on real LiDAR scans that contain raydrop, the Tweedie estimate $\hat{\bm{x}}_t$ also reflects plausible raydrop patterns based on the learned prior distribution (see~\cref{fig:model} for instance).
Therefore, we can construct a raydrop-aware mask $\bm{m}_t$ by thresholding the range 
channel of $\hat{\bm{x}}_t$, whose $i$-th element is determined as follows:
\begin{equation}
	\label{eq:masking}
	m_t^i =
	\begin{cases}
		1 & \text{if } \hat{x}_{t, \text{range}}^i > \eta \\
		0 & \text{otherwise}                              
	\end{cases},
\end{equation}
where $\hat{x}_{t, \text{range}}^i$ is the estimated range value of the $i$-th pixel at the timestep $t$, and $\eta$ is a threshold value.
Accordingly, the guidance enforces consistency with the synthetic input $\bm{y}$ only at valid-return pixels ($m_t^i = 1$), while raydrop pixels ($m_t^i = 0$) are left unaffected, allowing the diffusion prior to freely maintain the learned raydrop patterns.
As shown in~\cref{fig:ablation_masked_guidance}, our approach prevents over-inpainting and successfully reproduces realistic raydrop noise and reflectance intensity.

\myparagraph{Initialization.}
The state variable $\bm{x}_t$ of~\cref{eq:reverse_sde} evolves from Gaussian noise $\bm{x}_1$ during sampling.
Thus, the Tweedie sample $\hat{\bm{x}}_t$ and its derived $\bm{m}_t$ may not be aligned with the target simulation $\bm{y}$ in the early steps, leading to unstable progression.
To mitigate this issue, we initialize the low-frequency component of evolving sample $\bm{x}_t$ based on $\bm{y}$, following a strategy similar to SDEdit~\cite{meng2022sdedit}.
Specifically, we add noise to the simulation input $\bm{y}$ according to the forward SDE~\cite{song2021score-based} up to time $t_{\rm{init}}$.
The resulting noisy latent $\bm{x}_{t_{\rm{init}}}$ serves as the starting point of the reverse diffusion process.
We illustrate this step in~\cref{fig:model}(a).

\myparagraph{Harmonization.}
Applying masked guidance can create unnatural seams at the boundaries between the guided regions (where $m_t^i=1$) and the unconditionally generated regions (where $m_t^i=0$).
This is because the two areas are constrained by different dynamics. To harmonize these distinct areas, we employ the iterative resampling technique introduced in RePaint~\cite{lugmayr2022repaint}. 
For each timestep $t$ in the guided phase ($t < t_{\rm{init}}$), a single reverse step is replaced by a refinement loop of unconditional reverse and forward SDE steps. 
This iterative procedure allows information to propagate naturally between the two regions, ensuring that they are seamlessly blended into a coherent and realistic output.

\subsection{Architecture of Diffusion Prior Models}
For the prior term of~\cref{eq:masked_pigdm}, we build an unconditional diffusion model over 2-channel LiDAR range--reflectance images. 
The backbone is an encoder--decoder neural network (U-Net~\cite{ronneberger2015u-net}) conditioned on the timestep information; the network takes the state variable $\bm{x}_t$ and a log-SNR embedding as input.
We adopt the standard $\bm{\epsilon}$-prediction re-parameterization~\cite{ho2020denoising}; an involved Gaussian noise $\bm{\epsilon}$ is predicted by the neural network $\epsilon_\theta(\bm{x}_t, t)$, rather than directly estimating the score.
We denote the unconditional score (as in Sec.~\ref{sec:method}) by $s_\theta(\bm{x}_t,t)\equiv\nabla_{\bm{x}_t}\log p(\bm{x}_t)$.
Under the variance-preserving parameterization of SDE~\cite{song2021score-based} with $p(\bm{x}_t\mid\bm{x}_0)=\mathcal{N}(\bm{x}_t\mid\alpha_t\bm{x}_0,\sigma_t^2\bm{I})$, the score admits a closed-form conversion from $\bm{\epsilon}$-prediction:
\begin{equation}
	s_\theta(\bm{x}_t,t)=-\frac{\epsilon_\theta(\bm{x}_t,t)}{\sigma_t},
\end{equation}
and we can obtain the Tweedie sample $\hat{\bm{x}}_t$ by
\begin{equation}
	\hat{\bm{x}}_t = \frac{\bm{x}_t - \sigma_t\, \epsilon_\theta(\bm{x}_t, t)}{\alpha_t},
\end{equation}
where $\alpha_t$ and $\sigma_t$ are the noise schedule coefficients, which can be calculated from $f(\bm{x}_t,t)$ and $g(t)$ of~\cref{eq:reverse_sde}.
For numerical stability, we clip $\hat{\bm{x}}_t$ to a fixed range.

\subsection{Training Segmentation Models}

Finally we train segmentation models with the generated pseudo-real samples.
From the aforementioned posterior sampling, we obtain the sample $\bm{x}_0$ translated from $\bm{y}$.
To ensure consistency with the annotated labels of the simulation input $\bm{y}$, we re-initialize the valid range values of $\bm{x}_0$ with $\bm{y}$ while keeping the generated raydrop noise.
Similar to~\cref{eq:masking}, we first extract a raydrop mask $\bm{m}_0$ from $\bm{x}_{0, \text{range}}$.
We then apply $\bm{m}_0$ to the synthetic range $\bm{y}_{\text{range}}$ and the generated reflectance $\bm{x}_{0, \text{reflect}}$.
Both are then concatenated for training segmentation models.
For training point-based and voxel-based backbones~\cite{hu2020randla-net,choy20194d}, we convert the pseudo-real sample into 3D point clouds and sparse voxels, respectively.

\section{Experiments}
\label{sec:experiments}

In this section, we evaluate our approach in terms of sample fidelity (\cref{sec:sample_fidelity}) and Sim2Real performance of semantic segmentation (\cref{sec:semseg}).

\subsection{Settings}

\myparagraph{Datasets and tasks.}
The datasets used in our experiments are summarized in~\cref{tab:datasets}.
For the \emph{sample quality} evaluation, we use the SynLiDAR dataset~\cite{xiao2022transfer} as a simulation dataset and SemanticKITTI dataset~\cite{behley2019semantickitti} as a real dataset.
We first train a diffusion model on the real dataset, translate the simulation dataset into the pseudo-real dataset, and then evaluate the distributional similarity between the real and pseudo-real samples.
For the \emph{semantic segmentation} evaluation, we conduct two scenarios: 
(1) Following Wu \etal~\cite{wu2019squeezesegv2}, we train 2-class segmentation models on the GTA-LiDAR dataset~\cite{wu2019squeezesegv2} and evaluate on the 90° frontal subset~\cite{wu2019squeezesegv2} of KITTI dataset~\cite{geiger2013vision}, referred to as KITTI-frontal. 
(2) We also conduct 19-class semantic segmentation. We use SynLiDAR dataset~\cite{xiao2022transfer} for training and SemanticKITTI dataset~\cite{behley2019semantickitti} for evaluation on the common 19 classes.

\begin{table}[tb]
	\centering
	\footnotesize
	\caption{Two scenarios in our experiments}
	\label{tab:datasets}
	\begin{threeparttable}
		\begin{tabularx}{\hsize}{Lcccc}
			\toprule
			Dataset                                      & Domain     & \#Classes          & \#Samples & Resolution                      \\
			\midrule
			GTA-LiDAR~\cite{wu2019squeezesegv2}          & Simulation & \multirow{2}{*}{2} & 121,087   & \multirow{2}{*}{64$\times$512}  \\
			KITTI Raw~\cite{geiger2013vision}            & Real       &                    & \s10,848  &                                 \\
			\midrule
			SynLiDAR~\cite{xiao2022transfer}             & Simulation & 32$^\dagger$       & 198,396   & \multirow{2}{*}{64$\times$1024} \\
			SemanticKITTI~\cite{behley2019semantickitti} & Real       & 25$^\dagger$       & \s43,552  &                                 \\
			\bottomrule
		\end{tabularx}
		\begin{tablenotes}
			\item $\dagger$~We use the common 19 classes as listed in~\cref{tab:quantitative_19class_segmentation}.
		\end{tablenotes}
	\end{threeparttable}
\end{table}

\myparagraph{Implementation details.}
For building diffusion priors of real data, we employ R2DM~\cite{nakashima2024lidar}, a continuous-time diffusion model of range/reflectance images.
To improve generation quality and fidelity, we increase the model capacity of the official implementation\footnote{\url{https://github.com/kazuto1011/r2dm}} from 31M to 285M parameters by doubling the number of channels.
Pseudo-real samples are generated via posterior sampling as in Sec.~\ref{sec:method}.
We set $t_{\mathrm{init}}=0.8$ for initialization.
Similar to $\Pi$GDM~\cite{song2023pseudoinverse-guided}, we use DDIM sampler~\cite{song2021denoising} for~\cref{eq:reverse_sde} with discretization of 32 uniform steps, each with our masked guidance of~\cref{eq:masked_pigdm}. 
The progressive mask $\bm{m}_t$ is built from the Tweedie estimate by thresholding the normalized range in $[-1,1]$ using $\eta=-0.3$.
For the harmonization process, we perform three additional resampling cycles for each sampling step.
The entire pipeline requires approximately 8 seconds per sample on a single NVIDIA RTX 6000 Ada GPU.

\subsection{Sample Fidelity}
\label{sec:sample_fidelity}

\myparagraph{Baselines.}
We compare four methods for simulating reflectance intensity and raydrop noise, as listed in \cref{tab:quantitative_fidelity}.
The rendering model in~\cref{tab:quantitative_fidelity} represents the officially-provided SynLiDAR samples\footnote{\url{https://github.com/xiaoaoran/SynLiDAR}}, where the reflectance intensity is generated using the voxel-based supervised model~\cite{xiao2022transfer}.
For diffusion-based approaches, we compare three methods using different components of our framework:
SDEdit~\cite{meng2022sdedit} which initializes the sampling at $t_{\rm{init}}$,
$\Pi$GDM~\cite{song2023pseudoinverse-guided} which uses the original pseudoinverse guidance without our progressive mask $\bm{m}_t$,
and our DRUM which integrates all components.
All diffusion methods use the same unconditional R2DM model pre-trained on SemanticKITTI~\cite{behley2019semantickitti}.

\myparagraph{Evaluation metrics.}
To evaluate the sample fidelity, we compute the distributional similarity between real and pseudo-real samples across multiple levels of data representation. 
We use five evaluation metrics that extend the Fr\'echet Inception distance (FID) for LiDAR data: Fr\'echet range distance (FRD)~\cite{zyrianov2022learning}, Fr\'echet range image distance (FRID)~\cite{ran2024towards}, Fr\'echet point cloud distance (FPD)~\cite{shu20193d}, Fr\'echet point-based volume distance (FPVD)~\cite{ran2024towards}, and Fr\'echet sparse volume distance (FSVD)~\cite{ran2024towards}.
FRD and FRID evaluate similarity in range image format, FPD in point cloud format, and FPVD and FSVD in voxel format.
FRD uses both range and reflectance channels for evaluation, while FRID focuses solely on the range channel.
The point cloud and voxel representations are derived from the range image through spherical projection and voxelization, respectively.
\cref{tab:quantitative_fidelity} summarizes the correspondence between metrics and data representations.

\myparagraph{Quantitative results.}
\cref{tab:quantitative_fidelity} presents the quantitative results of the fidelity evaluation.
Our proposed method significantly improves the fidelity of SynLiDAR samples~\cite{xiao2022transfer} constructed by the simulator. 
However, our scores are close to those of SDEdit, which only sets the initial value of the reverse diffusion process. 
We attribute these results to the limitations of the evaluation, since fidelity metrics quantify distributional similarity, not per-sample quality. 
We further discuss the difference between our method and SDEdit in the qualitative results.

\begin{table*}[t]
	\definecolor{depth}{rgb}{0.282,0.471,0.816}
	\definecolor{rflct}{rgb}{0.933,0.522,0.290}
	\definecolor{point}{rgb}{0.416,0.800,0.392}
	\definecolor{voxel}{rgb}{0.839,0.373,0.373}
	\definecolor{bev}{rgb}{0.584,0.424,0.706}
	\newcommand{\best}[1]{\cellcolor{gray!20}\textbf{#1}}
	\newcommand{\subopt}[1]{\cellcolor{gray!20}#1}
	\centering
	\footnotesize
	\begin{threeparttable}
		\caption{Quantitative comparison of Sim2Real methods on sample fidelity.}
		\label{tab:quantitative_fidelity}
		\begin{tabularx}{\hsize}{l cc CC C CC}
			\toprule
			Method   & Raydrop   & Reflectance                   & 
			\twocolorsquare{depth}{rflct}~FRD~ $\downarrow$ & 
			\twocolorsquare{depth}{depth}~FRID~ $\downarrow$ & 
			\twocolorsquare{point}{point}~FPD~$\downarrow$ & 
			\twocolorsquare{point}{voxel}~FPVD $\downarrow$ &
			\twocolorsquare{voxel}{voxel}~FSVD $\downarrow$ \\
			\midrule
			Rendering model~\cite{xiao2022transfer} &            & \checkmark & 2198.8         & 76.6          & 148.1          & 60.3          & 69.9          \\
			\midrule
			SDEdit~\cite{meng2022sdedit}            & \checkmark & \checkmark & \best{867.6}   & \subopt{40.2} & \subopt{140.1} & \subopt{55.1} & \subopt{63.4} \\
			$\Pi$GDM~\cite{song2023pseudoinverse-guided}
			                                        & \checkmark & \checkmark & 1413.9         & 75.0          & 146.1          & 59.8          & 69.4          \\
			\textbf{DRUM} (ours)                    & \checkmark & \checkmark & \subopt{921.0} & \best{39.6}   & \best{139.4}   & \best{54.2}   & \best{62.5}   \\
			\bottomrule
		\end{tabularx}
		\begin{tablenotes}
			\item Notation: \twocolorsquare{depth}{depth} range image, \twocolorsquare{rflct}{rflct} reflectance image, \twocolorsquare{point}{point} point cloud, \twocolorsquare{voxel}{voxel} voxel. The best score is in \textbf{bold} and the top two scores are \colorbox{gray!20}{shaded}.
		\end{tablenotes}
	\end{threeparttable}
\end{table*}

\begin{figure*}[t]
	\footnotesize
	\centering
	\newcommand{\cmarkc}{\textcolor{green}{\cmark}}
	\newcommand{\xmarkc}{\textcolor{red}{\xmark}}
	\setlength{\tabcolsep}{2pt}
	\begin{tabularx}{\hsize}{C|CCC|C}
		Simulation~\cite{xiao2022transfer} (input) & SDEdit~\cite{meng2022sdedit}          & $\Pi$GDM~\cite{song2023pseudoinverse-guided} & \textbf{DRUM} (ours)                  & Real~\cite{behley2019semantickitti} (reference) \\
		\includegraphics[width=\hsize,height=0.08\textheight,keepaspectratio,trim=0 4 0 4,clip]{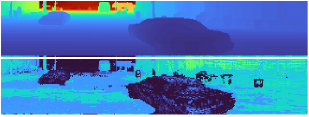} &   
		\includegraphics[width=\hsize,height=0.08\textheight,keepaspectratio,trim=0 4 0 4,clip]{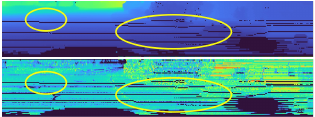}   &   
		\includegraphics[width=\hsize,height=0.08\textheight,keepaspectratio,trim=0 4 0 4,clip]{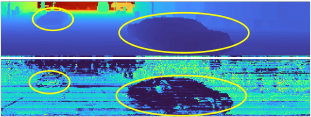}    &   
		\includegraphics[width=\hsize,height=0.08\textheight,keepaspectratio,trim=0 4 0 4,clip]{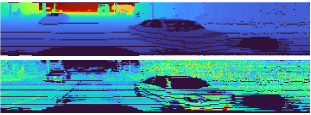}     &   
		\includegraphics[width=\hsize,height=0.08\textheight,keepaspectratio,trim=0 4 0 4,clip]{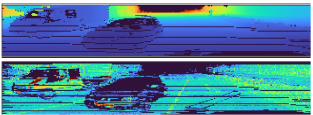} \\
		                                           & \xmarkc~Consistency~~\cmarkc~Fidelity & \cmarkc~Consistency~~\xmarkc~Fidelity        & \cmarkc~Consistency~~\cmarkc~Fidelity &                                                 
	\end{tabularx}
	\caption{\textbf{Qualitative comparison of pseudo-real samples.} We compare the range (top row) and reflectance (bottom row) samples produced by different methods. The reflectance input is from the rendering model~\cite{xiao2022transfer} while we do not use it for producing the pseudo-real samples. Our method shows better results in terms of consistency to the input and fidelity comparable to the reference.}
	\label{fig:qualitative_fidelity}
\end{figure*}

\myparagraph{Qualitative results.}
In \cref{fig:qualitative_fidelity}, we compare the pseudo-real samples generated by different methods. 
SDEdit alone generates realistic reflectance intensity and raydrop noise, but fails to preserve the scene content from the simulation data, such as vehicles and the cityscape, as highlighted by the yellow circles.
Conversely, $\Pi$GDM alone preserves the scene content but fails to reproduce raydrop noise due to the strong guidance from the likelihood term.
In contrast, our method successfully generates realistic reflectance intensity and raydrop noise while maintaining high fidelity to the original scene content.
Notably, our approach accurately reproduces the raydrop noise around car windows, which can be seen frequently in real samples due to the reflective properties of glass.

\subsection{Semantic Segmentation Results}
\label{sec:semseg}

\myparagraph{Baselines.}
For the first scenario, we follow the protocol of prior work~\cite{nakashima2023generative} to evaluate different methods of rendering raydrop noise, as listed in Table~\ref{tab:quantitative_2class_segmentation}. 
We compare against several methods: no rendering (config-A), global statistics (config-B), spatial statistics~\cite{wu2019squeezesegv2} (config-C), ePointDA~\cite{zhao2021epointda} (config-D), DUSty~\cite{nakashima2021learning} (config-E), and DUSty v2~\cite{nakashima2023generative} (config-F). 
The global/spatial statistics methods are Bernoulli sampling based on the average ratio of raydrop pixels.
The DUSty methods are based on the raydrop probability map generated by GANs.
ePointDA is an unpaired mapping model trained to predict the raydrop noise using CycleGAN~\cite{zhu2017unpaired}.
Our proposed diffusion-based approach is denoted as config-G.
We use SqueezeSegV2~\cite{wu2019squeezesegv2} for the semantic segmentation architecture across all configurations.
In our second scenario, we evaluate the effectiveness of our approach across different representations of LiDAR data on the 19-class semantic segmentation task. 
We train segmentation models on range images, point clouds, and voxels.
We employ RangeNet~\cite{milioto2019rangenet++}, RandLA-Net~\cite{hu2020randla-net}, and MinkowskiNet~\cite{choy20194d}, respectively.

\begin{table}[t]
	\newcommand{\best}[1]{\cellcolor{gray!20}\textbf{#1}}
	\newcommand{\subopt}[1]{\cellcolor{gray!20}#1}
	\begin{threeparttable}
		\caption{Quantitative comparison of Sim2Real methods on semantic segmentation (GTA-LiDAR $\rightarrow$ KITTI-frontal)}
		\label{tab:quantitative_2class_segmentation}
		\begin{tabularx}{\hsize}{clCcC}
			\toprule
			&                                              & \multicolumn{3}{c}{IoU (\%, $\uparrow$)} \\
			\cmidrule{3-5}
			Config & Method                                       & Car           & Pedestrian    & Mean          \\
			\midrule
			A      & No rendering                                 & \s1.1         & \s2.4         & \s1.7         \\
			B      & Global statistics                            & 55.2          & 25.1          & 40.2          \\
			C      & Spatial statistics~\cite{wu2019squeezesegv2} & 59.0          & 22.5          & 40.7          \\
			\midrule
			D      & ePointDA~\cite{zhao2021epointda}             & 66.2          & 24.8          & 45.5          \\
			E      & DUSty~\cite{nakashima2021learning}           & 59.1          & \best{28.0}   & 43.5          \\
			F      & DUSty v2~\cite{nakashima2023generative}      & \best{67.3}   & 25.2          & \subopt{46.3} \\
			G      & \textbf{DRUM} (ours)                         & \subopt{66.5} & \subopt{26.9} & \best{46.7}   \\
			\bottomrule
		\end{tabularx}
	\end{threeparttable}
\end{table}

\myparagraph{Quantitative results.}
The results for the two experimental scenarios are presented in \cref{tab:quantitative_2class_segmentation,tab:quantitative_19class_segmentation}, respectively.
In the first 2-class scenario (\cref{tab:quantitative_2class_segmentation}), our diffusion-based approach (config-G) achieves the best performance with an mIoU of 46.7\%, surpassing all baseline methods.
In the second 19-class scenario (\cref{tab:quantitative_19class_segmentation}), our method consistently improves performance across all data representations.
A key factor in this success is the realistic modeling of raydrop.
Compared to the baseline rendering, our DRUM model using only raydrop yields substantial gains across all representations, with image-based models showing a more than a twofold increase in mIoU.
This improvement is particularly notable in the car class. As shown in our fidelity evaluation (\cref{fig:qualitative_fidelity}), our method successfully reproduces the distinct raydrop patterns around car windows, which frequently arise in real LiDAR scans due to reflections from glass surfaces. 
In contrast, the contribution of reflectance is more nuanced.
We hypothesize that subtle inconsistencies between the range and reflectance modalities can offset the potential benefits of multimodality.
Addressing this issue to better exploit both modalities remains an important direction for future work.

\begin{table*}[t]
	\centering
	\scriptsize
	\newcommand{\best}[1]{\cellcolor{gray!20}\textbf{#1}}
	\newcommand{\subopt}[1]{\cellcolor{gray!20}#1}
	\setlength{\tabcolsep}{2.2pt}
	\caption{Quantitative comparison of Sim2Real methods on semantic segmentation (SynLiDAR $\rightarrow$ SemanticKITTI).}
	\label{tab:quantitative_19class_segmentation}
	\begin{tabularx}{\hsize}{llccCCCCCCCCCCCCCCCCCCCCC}
		\toprule
		& & & & \multicolumn{20}{c}{Intersection-over-Union (IoU, \%) $\uparrow$} \\
		\cmidrule(lr){5-24}
		Network & Sim2Real method &
		\rotatebox{90}{Raydrop} &
		\rotatebox{90}{Reflectance} &
		\rotatebox{90}{\mysquare[0.39, 0.59, 0.96] Car} &
		\rotatebox{90}{\mysquare[0.39, 0.90, 0.96] Bicycle} &
		\rotatebox{90}{\mysquare[0.12, 0.24, 0.59] Motorcycle} &
		\rotatebox{90}{\mysquare[0.31, 0.12, 0.71] Truck} &
		\rotatebox{90}{\mysquare[0.00, 0.00, 1.00] Bus} &
		\rotatebox{90}{\mysquare[1.00, 0.12, 0.12] Person} &
		\rotatebox{90}{\mysquare[1.00, 0.16, 0.78] Bicyclist} &
		\rotatebox{90}{\mysquare[0.59, 0.12, 0.35] Motorcyclist} &
		\rotatebox{90}{\mysquare[1.00, 0.00, 1.00] Road} &
		\rotatebox{90}{\mysquare[1.00, 0.59, 1.00] Parking} &
		\rotatebox{90}{\mysquare[0.60, 0.00, 0.71] Sidewalk} &
		\rotatebox{90}{\mysquare[0.69, 0.00, 0.29] Other-ground} &
		\rotatebox{90}{\mysquare[1.00, 0.78, 0.00] Building} &
		\rotatebox{90}{\mysquare[1.00, 0.47, 0.20] Fence} &
		\rotatebox{90}{\mysquare[0.00, 0.69, 0.00] Vegetation} &
		\rotatebox{90}{\mysquare[0.53, 0.24, 0.00] Trunk} &
		\rotatebox{90}{\mysquare[0.59, 0.94, 0.31] Terrain} &
		\rotatebox{90}{\mysquare[1.00, 0.94, 0.59] Pole} &
		\rotatebox{90}{\mysquare[1.00, 0.00, 0.00] Traffic-sign} &
		\rotatebox{90}{Mean (mIoU)} \\
		\midrule
		\multirow{4}{*}{Image-based~\cite{milioto2019rangenet++}} & – &  &  &
		\s6.2         & \s2.0          & \s0.4          & \s2.1          & \s1.1          & \s3.3          & \s5.8         & \s0.0          & \s8.0         & \s0.5          & \s6.5         & \s0.0        & 28.6          & \s2.3         & \best{49.0}   & 13.7          & 20.3          & 12.9          & \best{\s2.2}   & \s8.7         \\
		& Rendering model~\cite{xiao2022transfer} &  & \checkmark &
		\s3.4         & \best{\s3.1}   & \s0.4          & \best{\s4.9}   & \s1.4          & \s3.0          & \s3.3         & \s0.0          & \s5.8         & \s0.2          & \s8.3         & \s0.0        & 30.8          & \s3.6         & 33.9          & 15.3          & \subopt{21.6} & 12.7          & \s1.5          & \s8.1         \\
		& \textbf{DRUM} (ours) & \checkmark     &                &
		\subopt{56.2} & \subopt{\s2.7} & \subopt{\s5.9} & \subopt{\s3.8} & \best{\s7.1}   & \best{\s6.7}   & \best{19.3}   & \s0.0          & \subopt{49.3} & \best{\s7.4}   & \subopt{30.2} & \s0.0        & \subopt{35.2} & \best{\s4.6}  & 31.7          & \best{18.9}   & 19.3          & \best{17.5}   & \subopt{\s2.1} & \subopt{16.7} \\
		& \textbf{DRUM} (ours) & \checkmark     & \checkmark     &
		\best{57.3}   & \s1.9          & \best{\s8.6}   & \s3.1          & \subopt{\s5.5} & \subopt{\s5.8} & \subopt{14.0} & \s0.0          & \best{52.8}   & \subopt{\s5.9} & \best{32.6}   & \s0.0        & \best{40.3}   & \best{\s4.1}  & \subopt{40.7} & \subopt{18.6} & \best{25.3}   & \subopt{14.6} & \s1.9          & \best{17.5}   \\
		\midrule
		\multirow{4}{*}{Point-based~\cite{hu2020randla-net}} & – &  &  &
		29.6          & \s5.8          & \s2.1          & \subopt{\s1.4} & \s3.4          & 13.0           & 29.0          & \s0.1          & 12.3          & \s2.1          & 30.6          & \s0.0        & 37.1          & \s5.5         & 61.0          & 23.1          & \subopt{38.0} & 35.5          & \subopt{12.0}  & 18.0          \\
		& Rendering model~\cite{xiao2022transfer} &  & \checkmark &
		32.2          & \s1.2          & \best{11.6}    & \best{\s2.1}   & \best{\s4.3}   & 16.1           & 31.5          & \s0.0          & \s9.3         & \s4.3          & 28.7          & \s0.0        & 54.7          & \s3.8         & \best{65.8}   & 14.2          & 15.9          & 19.2          & \s8.0          & 17.0          \\
		& \textbf{DRUM} (ours) & \checkmark    &                &
		\subopt{40.5} & \best{16.4}    & \subopt{6.6}   & 0.9            & 0.6            & \subopt{18.0}  & \best{51.1}   & \subopt{0.3}   & \best{23.0}   & \subopt{9.6}   & \best{32.1}   & 0.0          & \subopt{58.5} & \subopt{10.3} & \subopt{64.1} & \subopt{24.6} & \best{38.6}   & \subopt{38.5} & 8.8            & \subopt{23.3} \\
		& \textbf{DRUM} (ours) & \checkmark    & \checkmark     &
		\best{57.4}   & \subopt{6.6}   & 5.9            & 1.0            & \best{12.0}    & \best{19.3}    & \subopt{36.1} & \best{0.8}     & \subopt{16.8} & \best{12.5}    & \subopt{30.9} & 0.0          & \best{61.7}   & \best{17.9}   & 61.4          & \best{30.5}   & 33.0          & \best{40.6}   & \best{12.1}    & \best{24.0}   \\
		\midrule
		\multirow{4}{*}{Voxel-based~\cite{choy20194d}} & – & & &
		55.5          & 10.3           & 25.5           & \subopt{\s1.4} & \s2.7          & 17.6           & 50.7          & \s2.8          & 12.5          & \best{\s7.6}   & 32.3          & \s0.0        & 29.9          & \best{19.0}   & 62.4          & 21.1          & 37.7          & 11.5          & \s1.9          & 21.2          \\
		& Rendering model~\cite{xiao2022transfer} &  & \checkmark &
		61.0          & \s8.3          & \subopt{28.9}  & \s\subopt{1.4} & \s2.5          & 16.0           & 53.2          & \s1.7          & 15.8          & \s4.2          & 32.0          & \s0.0        & 30.5          & \s9.1         & 62.4          & 16.5          & 35.0          & \s7.4         & \s2.2          & 20.4          \\
		& \textbf{DRUM} (ours) & \checkmark    &                &
		\subopt{73.1} & \subopt{20.2}  & 28.5           & \best{\s2.4}   & \subopt{\s4.9} & \subopt{20.2}  & \subopt{55.5} & \best{\s5.8}   & \best{17.7}   & \subopt{\s7.0} & \best{34.8}   & \best{\s0.1} & \subopt{47.2} & \subopt{13.6} & \subopt{67.9} & \subopt{29.4} & \best{52.8}   & \subopt{23.5} & \subopt{\s4.4} & \subopt{26.8} \\
		& \textbf{DRUM} (ours)  & \checkmark    & \checkmark     &
		\best{76.3}   & \best{23.3}    & \best{33.2}    & \s1.3          & \best{11.4}    & \best{31.1}    & \best{57.2}   & \subopt{\s5.6} & \subopt{16.6} & \s\subopt{7.0} & \subopt{33.9} & \s0.0        & \best{59.1}   & \s9.4         & \best{72.0}   & \best{34.3}   & \subopt{52.0} & \best{29.7}   & \best{\s5.9}   & \best{29.4}   \\
		\bottomrule
	\end{tabularx}
\end{table*}

\begin{figure*}[t]
	\centering
	\footnotesize
	\setlength{\tabcolsep}{3pt}
	\begin{tabularx}{\hsize}{CCCC}
		Input & Ground truth & SynLiDAR (baseline) & \textbf{DRUM} (ours) \\
		\includegraphics[width=\hsize]{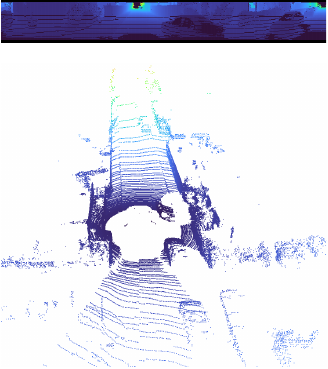} &
		\includegraphics[width=\hsize]{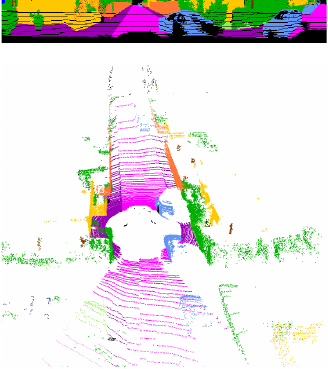} &
		\includegraphics[width=\hsize]{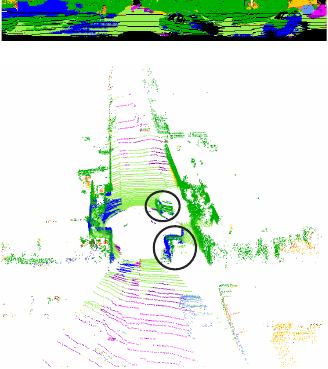} &
		\includegraphics[width=\hsize]{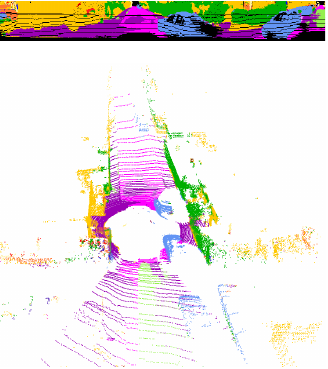} \\
	\end{tabularx}
	\caption{\textbf{Qualitative comparison of semantic segmentation results.} We show the results of the image-based method on the representation of images (top row) and point clouds (bottom row). Our approach significantly improves \mysquare[0.39, 0.59, 0.96] car detection (black circles) compared to the baseline.}
	\label{fig:qualitative_19class_segmentation}
\end{figure*}

\myparagraph{Qualitative results.}
\cref{fig:qualitative_19class_segmentation} presents a qualitative comparison of image-based segmentation results on the SemanticKITTI validation set.
Compared to the baseline model trained on the official SynLiDAR samples~\cite{xiao2022transfer}, our method produces results that are visually closer to the ground truth.
In particular, background classes such as \emph{vegetation} and \emph{building} are more consistently recognized, reducing the spurious regions observed in the baseline.
For foreground objects, the baseline often struggles to disentangle cars from their surroundings, resulting in blurred boundaries and cases where cars are partially merged with vegetation or nearby structures.
By contrast, our method better preserves \emph{cars}, yielding clearer object boundaries and fewer false negatives.

\begin{figure}[t]
	\centering
	\includegraphics[width=0.9\hsize]{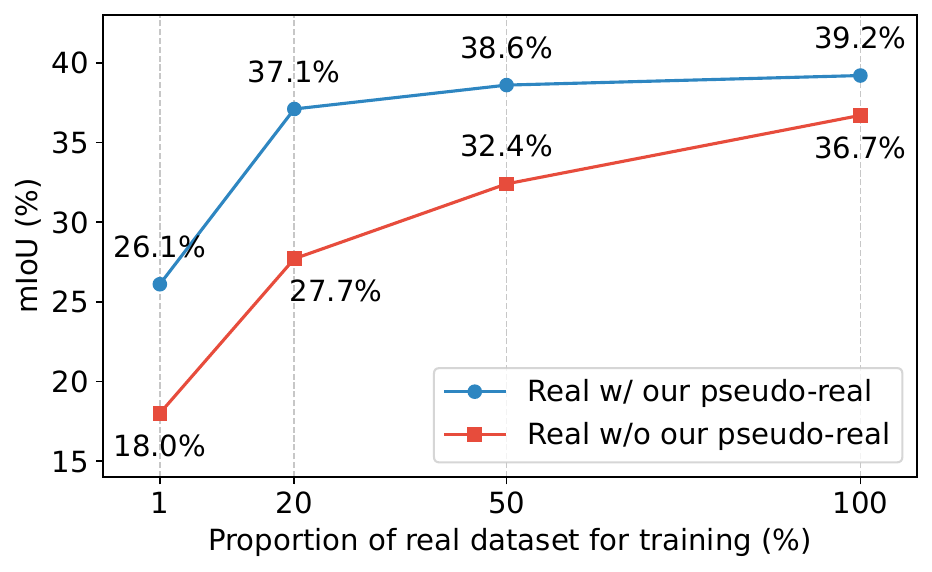}
	\caption{\textbf{Effect of mixed training using our pseudo-real samples.} We show scores of mIoU (\%) on image-based 19-class semantic segmentation. We mix real data at ratios of 1\%, 20\%, 50\%, and 100\% with our pseudo-real data during training.}
	\label{fig:quantitative_mixed_training}
\end{figure}

\begin{figure}[t]
	\centering
	\footnotesize
	\setlength{\tabcolsep}{3pt}
	\begin{tabularx}{\hsize}{C|CC}
		\includegraphics[width=\hsize]{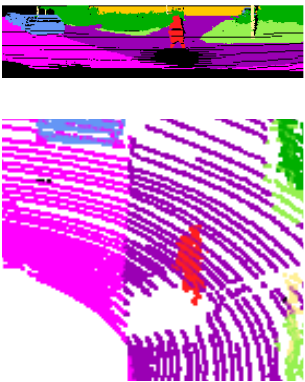} &
		\includegraphics[width=\hsize]{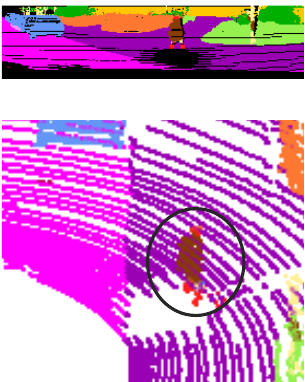} &
		\includegraphics[width=\hsize]{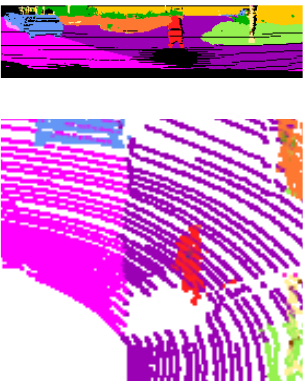} \\
		Ground truth & Real only & Real + our pseudo-real 
	\end{tabularx}
	\caption{\textbf{Qualitative comparison of segmentation results in the mixed training scenario.} When trained solely on real data (middle), the model misclassifies the \mysquare[1.00, 0.12, 0.12] person as the \mysquare[0.53, 0.24, 0.00] trunk.}
	\label{fig:qualitative_mixed_training}
\end{figure}

\myparagraph{Mixed training.}
Given the notable improvements achieved by our method on image-based 19-class semantic segmentation, we further evaluate a mixed training scenario with a limited labeled real dataset.
To assess the impact of different levels of data availability, we systematically sample labeled subsets from SemanticKITTI at ratios of 1\%, 20\%, 50\%, and 100\%.
We then train RangeNet~\cite{milioto2019rangenet++} using mini-batches that are equally sampled from real and pseudo-real data.
The results are shown in \cref{fig:quantitative_mixed_training}.
Training with real data alone achieves a validation mIoU of 36.7\%.
Using only 20\% of the real data combined with pseudo-real samples improves the mIoU to 37.1\%, already surpassing the real-only baseline.
When all real data are combined with pseudo-real samples, the mIoU further increases to 39.2\%.
Notably, we observe a substantial improvement in minority classes.
While real-only training achieves an IoU of 5.2\% for the person class, mixed training improves this to 20.5\%.
An IoU of the bicyclist class was also improved from 28.8\% to 51.1\%.
This improvement is clearly reflected in the qualitative results shown in \cref{fig:qualitative_mixed_training}.

\section{Conclusions}
In this paper, we introduced DRUM, a novel Sim2Real translation framework for LiDAR semantic segmentation. By formulating domain adaptation as posterior sampling using diffusion models, DRUM generates realistic pseudo-real data that capture complex measurement characteristics such as reflectance intensity and raydrop noise. Our raydrop-aware masked guidance mechanism preserves geometric consistency while enabling faithful translation of domain-specific features. Our experiments demonstrated that DRUM outperforms existing methods in terms of sample fidelity and consistently improves semantic segmentation performance across image, point cloud, and voxel representations. In future work, we plan to explore emerging flow matching models as potential alternatives to diffusion priors, aiming for faster and more flexible Sim2Real adaptation.

\bibliographystyle{ieeetr}
\bibliography{main}

\end{document}